\documentclass[conference]{IEEEtran}
\IEEEoverridecommandlockouts

\usepackage{cite}
\usepackage{amsmath,amssymb,amsfonts}
\usepackage{graphicx}
\usepackage{booktabs}
\usepackage{multirow}
\usepackage{array}
\usepackage{tabularx}
\usepackage{url}
\usepackage[colorlinks=true, linkcolor=blue, urlcolor=blue, citecolor=blue]{hyperref}
\usepackage{subcaption}
\usepackage{xcolor}
\usepackage{algorithmic}
\usepackage{algorithm}
\usepackage{balance}

\def\BibTeX{{\rm B\kern-.05em{\sc i\kern-.025em b}\kern-.08em
    T\kern-.1667em\lower.7ex\hbox{E}\kern-.125emX}}

\begin{document}

\title{PhaseNet++: Phase-Aware Frequency-Domain Anomaly Detection\\
for Industrial Control Systems via Phase Coherence Graphs}

\author{
\IEEEauthorblockN{1\textsuperscript{nd} Raviteja Bommireddy}
\IEEEauthorblockA{\textit{Department of CSE} \\
\textit{IIITDM Kancheepuram}\\
Chennai, India\\
ravitejab1978@gmail.com}
\and
\IEEEauthorblockN{2\textsuperscript{nd} Varshith Bandaru}
\IEEEauthorblockA{\textit{Department of CSE} \\
\textit{IIITDM Kancheepuram}\\
Chennai, India\\
varshithb30@gmail.com}
\and
\IEEEauthorblockN{3\textsuperscript{rd} Lohith Pakala}
\IEEEauthorblockA{\textit{Department of CSE} \\
\textit{IIITDM Kancheepuram}\\
Chennai, India\\
lohith.pakala.i@gmail.com}
\and
\IEEEauthorblockN{4\textsuperscript{th} PradeepKumar B}
\IEEEauthorblockA{\textit{Department of CSE} \\
\textit{IIITDM Kancheepuram}\\
Chennai, India\\
pradeepkumar@iiitdm.ac.in}
}

\maketitle

\begin{abstract}
Multivariate time series anomaly detection in industrial control systems
(ICS) has attracted growing attention due to the increasing threat of
cyber-physical attacks on critical infrastructure. State-of-the-art
methods model inter-sensor relationships from raw time-domain
amplitude values, using graph neural networks, Transformers, or
variational autoencoders. However, these methods discard the
\emph{phase spectrum} produced by time frequency transformations, a
signal that directly encodes timing and synchronization relationships
between physically coupled sensors. We argue that phase information
constitutes a complementary and previously overlooked detection
modality for ICS anomaly detection.

We present \textbf{PhaseNet++}, a frequency-domain autoencoder that
operates on the Short-Time Fourier Transform (STFT) of sliding sensor
windows, retaining both magnitude and phase spectra. A \emph{Phase
Coherence Index} (PCI), inspired by the Phase Locking Value from
neuroscience, summarizes pairwise phase consistency across frequency
bins into a continuous adjacency matrix. This matrix guides a graph
attention network that propagates information preferentially among
phase-synchronized sensors. A sensor-token Transformer encoder captures
system-wide structure, and a dual-head decoder reconstructs magnitude
and phase jointly via circular and coherence-aware objectives. Evaluated on the Secure Water Treatment (SWaT) benchmark, PhaseNet++
achieves an F1-score of 90.98\%, ROC-AUC of 95.66\%, and average
precision of 91.51\%. Ablation studies show that the phase-aware
front-end and PCI graph module together add only 264,816 parameters
(5.0\% of the total 5.29M), demonstrating that the phase inductive
bias is lightweight. While the absolute F1-score is second best than that of all recent raw-value methods evaluated under different protocols, we position this work as the first systematic study of phase-domain anomaly detection for ICS, opening a new direction for the field. Our codes are available at \href{https://github.com/raviteja-bommireddy/PhaseNet}{https://github.com/raviteja-bommireddy/PhaseNet}
\end{abstract}

\begin{IEEEkeywords}
industrial control systems, anomaly detection, phase-aware learning,
Short-Time Fourier Transform, graph attention networks, Transformer,
phase coherence, SWaT
\end{IEEEkeywords}

\section{Introduction}
\label{sec:intro}

Industrial control systems (ICS) serve as the operational foundation of
critical infrastructure, including water treatment facilities, power
grids, chemical plants, and transportation systems. These environments
combine digital control logic with physical processes through tightly
connected networks of sensors and actuators. Because of this close
integration, a cyber-physical attack affects more than data alone---it
can directly influence plant behavior, potentially creating safety
risks and significant economic consequences~\cite{mathur2016swat,tantawy2022elements,
macas2022survey}.

To address this challenge, researchers have proposed a wide range of
data-driven anomaly detection methods. Early studies mainly relied on
recurrent and convolutional neural networks applied directly to raw
sensor time-series
data~\cite{kravchik2018detecting,malhotra2016lstm,
munir2018deepant,filonov2016lstm}. Later work introduced graph-based
approaches such as GDN~\cite{deng2021gdn}, which learns inter-sensor
dependencies by constructing adjacency relationships and predicting
future sensor values. Transformer-based models, including Anomaly
Transformer~\cite{xu2022anomalytransformer} and
GTA~\cite{chen2021gta}, use self-attention mechanisms to capture
temporal dependencies and have reported strong detection performance.
Variational approaches such as OmniAnomaly~\cite{su2019omni}
and InterFusion~\cite{li2021interfusion} instead rely on latent-space
reconstruction probabilities for anomaly identification.

Despite their differences, these methods share one common assumption:
sensor behavior is modeled primarily in the \emph{time domain}, where
each channel is treated as a sequence of amplitude values. In this
setting, the Fourier phase spectrum---which carries information about
within-window timing offsets and synchronization across
sensors---is usually ignored or removed during preprocessing. We argue
that this leaves an important aspect of ICS behavior underexplored.

\textbf{Why does phase matter in ICS?}\; In industrial control
processes, physical coupling is reflected not only in similar value
patterns but also in consistent \emph{timing relationships}. For
example, when a pump starts operating in a water treatment plant,
downstream flow, pressure, and level sensors react after characteristic
delays, producing stable phase relationships in the frequency
domain~\cite{pikovsky2001synchronization}. If an attacker replays past
sensor values or introduces small timing shifts in actuator commands,
the amplitude profile may still appear realistic while these phase
relationships become inconsistent. Methods that rely only on magnitude
must detect such behavior indirectly, whereas a phase-aware approach
can capture it more directly.

\textbf{Our contribution.}\; We introduce PhaseNet++, a phase-aware
frequency-domain graph--Transformer autoencoder designed for ICS attack
detection. The model makes three specific contributions:

\begin{enumerate}
    \item \textbf{Phase as a first-class signal.} We transform each
    multivariate sensor window via STFT and retain both magnitude and
    phase spectra. The phase spectrum is reconstructed with a
    \emph{circular loss} that respects the angular geometry of phase
    measurements~\cite{mardia2000directional}, rather than treating
    phase as another Euclidean channel.

    \item \textbf{Phase Coherence Index (PCI) as graph structure.}
    Pairwise phase consistency across frequency bins is summarized
    into a continuous adjacency matrix inspired by the Phase Locking
    Value (PLV)~\cite{lachaux1999plv}. This PCI matrix injects a
    physics-informed spatial prior into graph attention, biasing
    message passing toward sensors that are phase-synchronized under
    normal operation.

    \item \textbf{Efficient phase inductive bias.} The spectral CNN
    embedding and PCI-weighted graph attention together account for
    only 264,816 parameters (5.0\% of the model). The added
    representation cost is modest, yet it opens a signal family that
    has not been explored in the ICS anomaly detection literature.
\end{enumerate}

We evaluate PhaseNet++ on the widely studied Secure Water Treatment
(SWaT) dataset~\cite{mathur2016swat,goh2016swat} and provide a
comprehensive comparison against 15 baselines reported in the GDN and
Anomaly Transformer papers. We are transparent about the fact that
different published works use different evaluation protocols
(point-level~vs.~window-level, with~vs.~without point adjustment);
accordingly, we report our results under a clearly defined protocol
and discuss comparability in detail.

Our primary aim is \emph{not} to claim a new state-of-the-art
F1-score---our protocol differs from prior work, and our absolute
numbers are lower than some prior results obtained under
point-adjustment regimes. Instead, we aim to demonstrate that
\emph{phase-domain features provide a meaningful complementary signal
for ICS anomaly detection}, and that encoding phase coherence as graph
structure is a principled way to exploit that signal. We view this
paper as laying the groundwork for a new direction in which future
work can combine phase-aware and amplitude-aware representations.

\section{Related Work}
\label{sec:related}

We organize prior work along four axes, identifying the gap that
PhaseNet++ fills.

\subsection{Time-Domain ICS Anomaly Detection}

The earliest deep-learning approaches to ICS anomaly detection
operated directly on raw sensor values.
Filonov~et~al.~\cite{filonov2016lstm} used LSTM networks for
predictive modeling, while
Kravchik~and~Shabtai~\cite{kravchik2018detecting} showed that 1D
CNNs could outperform recurrent architectures on the SWaT benchmark.
Malhotra~et~al.~\cite{malhotra2016lstm} proposed an LSTM-based
encoder--decoder and Munir~et~al.~\cite{munir2018deepant} introduced
DeepAnT for unsupervised prediction-based detection.
Wang~et~al.~\cite{wang2022cyberattacks} and
Al-Dhaheri~et~al.~\cite{aldhaheri2022swat} further studied diverse
ML-driven and model-based approaches specific to water treatment.
Ha~et~al.~\cite{ha2022explainable} combined LSTM autoencoders with
explainability techniques. All these methods are fundamentally
rooted in the time-domain amplitude of sensor readings.

\subsection{Graph-Based Sensor Modeling}

The realization that ICS sensors are not independent---but linked
through physical processes---motivated graph-based detectors.
GDN~\cite{deng2021gdn} is a seminal example: it learns an adjacency
graph from sensor embeddings and uses graph attention to forecast
each sensor's next value, flagging large prediction residuals.
MTGNN~\cite{wu2020mtgnn} alternates graph learning and temporal
convolution for spatio-temporal forecasting.
Yu~et~al.~\cite{yu2018stgcn} proposed spatio-temporal graph
convolutions for traffic, a framework later adapted to CPS.
GTA~\cite{chen2021gta} combines learned graphs with Transformer
temporal modeling for IoT anomaly detection.

A key observation is that in all these methods, graph edges are
learned from \emph{amplitude correlations} or via end-to-end
trainable embeddings. None uses phase relationships as the basis for
edge construction. Our PCI bypasses fully learned adjacency and
instead derives edges from a physically interpretable quantity:
pairwise phase coherence.

\subsection{Transformer-Based Time Series Anomaly Detection}

Transformers~\cite{vaswani2017attention} have recently advanced
anomaly detection in multivariate time series. Anomaly
Transformer~\cite{xu2022anomalytransformer} introduces
\emph{association discrepancy}---the divergence between a learned
prior-association (Gaussian kernel over time) and a series-association
(self-attention weights). The minimax strategy amplifies
normal--abnormal distinguishability, achieving strong results on
multiple benchmarks including SWaT.
Informer~\cite{zhou2021informer} improves efficiency via ProbSparse
attention for long-horizon forecasting.

These models have significantly advanced sequence modeling, but
their inputs remain raw time-domain values. None operates on the
frequency-domain spectrum, and none models phase explicitly.

\subsection{Frequency-Domain and Phase-Aware Methods}
\label{sec:related_freq}

Frequency-domain reasoning remains uncommon in multivariate anomaly
detection. Spectral residual methods~\cite{ren2019spectral} exploit
saliency in the frequency domain but do not preserve inter-signal
phase information. In the ICS literature, the closest prior work we
identified is TFANet~\cite{liu2024tfanet}, which transforms
industrial operation data into amplitude and phase and fuses time-
and frequency-domain features for \emph{supervised} classification.
Our setting differs in three important respects: (i)~PhaseNet++ is
trained \emph{unsupervised} on normal data only, (ii)~it explicitly
models pairwise phase coherence as graph structure, and (iii)~it
reconstructs phase with a circular objective rather than treating
phase as one more Euclidean feature channel.

\subsection{Phase Synchrony and Circular Statistics}

Phase synchronization is a classic tool in signal processing and
dynamical-systems theory. Stable phase locking between coupled
oscillators is a hallmark of
synchronization~\cite{pikovsky2001synchronization}. The Phase Locking
Value (PLV)~\cite{lachaux1999plv} measures the concentration of
relative phase on the unit circle and has been widely used in
neuroscience to quantify functional connectivity between brain
regions. Directional
statistics~\cite{mardia2000directional,boashash1992instantaneous}
provides the mathematical framework for comparing angular
variables. Our PCI adapts PLV to the ICS context, measuring
cross-sensor phase coherence in a single window's frequency
representation.

\textbf{Summary of the gap.}\; To the best of our knowledge, no prior
work on ICS anomaly detection: (a)~retains the Fourier phase spectrum
as a first-class detection feature, (b)~constructs sensor-graph
adjacency from pairwise phase coherence, or (c)~reconstructs phase
with a circular loss. PhaseNet++ addresses all three.

\section{Methodology}
\label{sec:method}

\begin{figure*}[t]
    \centering
    \includegraphics[width=\textwidth]{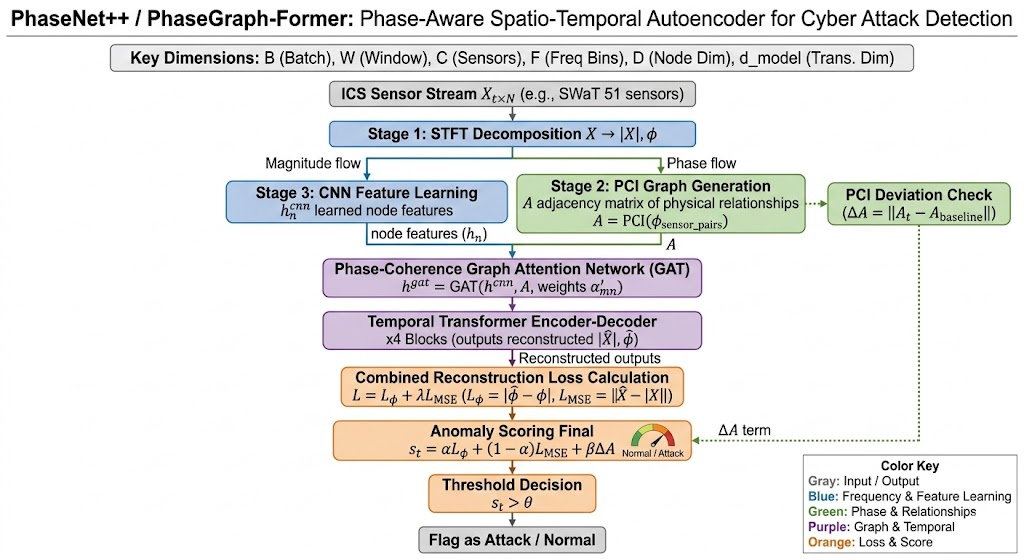}
    \caption{Overview of the PhaseNet++ pipeline. A multivariate
    sensor window $\mathbf{x}^{(t)} \in \mathbb{R}^{W \times C}$ is
    transformed by STFT into per-sensor magnitude and phase spectra.
    Pairwise phase differences define the PCI adjacency matrix~$\mathbf{A}$.
    A spectral CNN extracts per-sensor features, a PCI-weighted GAT
    propagates information across synchronized sensors, a sensor-token
    Transformer encoder captures system-wide structure, and a
    dual-head decoder reconstructs both magnitude and phase.}
    \label{fig:architecture}
\end{figure*}

\subsection{Problem Formulation}

Let $\mathbf{X} \in \mathbb{R}^{T \times C}$ denote a multivariate ICS
time series with $T$ time steps and $C$ sensors. Following the
standard unsupervised anomaly detection formulation
\cite{deng2021gdn,xu2022anomalytransformer}, we assume that only
normal data is available for training. PhaseNet++ operates on sliding
windows of length $W$:
\begin{equation}
    \mathbf{x}^{(t)} = \mathbf{X}_{t:t+W-1,:} \in \mathbb{R}^{W \times C}.
\end{equation}
At inference time, each window receives an anomaly score
$s^{(t)}$, and is classified as attack if $s^{(t)} > \tau$, where
$\tau$ is a threshold determined from the validation set.

\subsection{STFT Representation}
\label{sec:stft}

For each sensor $c \in \{1,\ldots,C\}$, we apply STFT with
$N_{\text{FFT}}=128$ to the windowed signal $x_c^{(t)}$. With window
length $W=100$ and no additional padding, each sensor yields a single
complex spectrum with $F = N_{\text{FFT}}/2 + 1 = 65$ frequency bins:
\begin{equation}
    z_c(f) = \sum_{n=0}^{W-1} x_c^{(t)}(n)\, g(n)\,
    e^{-j 2 \pi f n / N_{\text{FFT}}},
\end{equation}
where $g(\cdot)$ denotes the STFT analysis window. We decompose the
complex spectrum into magnitude and phase:
\begin{align}
    M_c(f) &= |z_c(f)|, \\
    \Phi_c(f) &= \angle z_c(f), \qquad \Phi_c(f) \in [-\pi,\pi].
\end{align}
Stacking all sensors yields the spectral tensor
$\mathbf{S} \in \mathbb{R}^{C \times 2 \times F}$,
whose two channels correspond to magnitude and phase, respectively.

\textbf{Design rationale.}\; Within a fixed window, magnitude captures
\emph{how strongly} each frequency component is excited, while phase
captures the \emph{within-window timing offset} of that component.
Existing ICS detectors implicitly compute magnitude-like features
(energy, variance) but systematically discard phase. We explicitly preserve $\Phi_c(f)$, viewing it as a source of timing
information that complements amplitude rather than duplicating it.
\subsection{Phase Coherence Index (PCI)}
\label{sec:pci}

To turn pairwise phase consistency into an explicit spatial prior, we
define the Phase Coherence Index between sensors $i$ and $j$ as:
\begin{equation}
    A_{ij} =
    \left|
    \frac{1}{F}
    \sum_{f=0}^{F-1}
    e^{j(\Phi_i(f)-\Phi_j(f))}
    \right|.
    \label{eq:pci}
\end{equation}
The resulting matrix $\mathbf{A} \in [0,1]^{C \times C}$ is symmetric,
contains ones along the diagonal, and assigns higher values to sensor
pairs whose relative phase remains concentrated across frequency bins.
Equation~\eqref{eq:pci} follows the idea of the
Phase Locking Value (PLV)~\cite{lachaux1999plv}, a measure commonly
used in neuroscience to estimate functional connectivity from EEG phase
signals. The main difference here is that PLV is adapted to a
\emph{single-window, multi-sensor} frequency representation instead of
the multi-trial setting typically used for brain recordings.

\textbf{Interpretation.}\; When two sensors are physically coupled and
respond to the same process with a stable lag during normal operation,
their relative phase remains consistent and $A_{ij}$ becomes high. If
an attack disrupts this coordination---for example, by replaying one
sensor's readings or introducing a slight actuation delay---the phase
difference becomes less concentrated across frequency bins, causing
$A_{ij}$ to decrease. In this way, PCI offers a
\emph{physics-guided} notion of adjacency that complements the
data-driven graph learning used in methods such as
GDN~\cite{deng2021gdn}.

\subsection{Network Architecture}

PhaseNet++ takes $(\mathbf{S}, \mathbf{A})$ as input and reconstructs
both magnitude and phase through four sequential stages.

\subsubsection{Spectral CNN Embedding}

Each sensor slice $\mathbf{S}_c \in \mathbb{R}^{2 \times F}$ is first
processed using a lightweight 1D CNN:
\begin{equation}
    \mathbf{h}^{(0)}_c = E(\mathbf{S}_c) \in \mathbb{R}^{D},
    \qquad D=128.
\end{equation}
The embedding network contains two convolutional layers (kernel
size 3, ReLU activations), followed by max-pooling and a linear
projection. At this stage, local spectral structures are extracted
jointly from magnitude and phase for each sensor independently.

\subsubsection{PCI-Weighted Graph Attention}

The resulting sensor embeddings are then refined through two layers of
graph attention (GAT)~\cite{velickovic2018gat}. For head $h$ in layer
$\ell$, the pre-softmax affinity is computed as:
\begin{equation}
    e_{ij}^{(\ell,h)} =
    \mathrm{LeakyReLU}
    \left(
        \frac{\mathbf{q}^{(\ell,h)}_i
        (\mathbf{k}^{(\ell,h)}_j)^\top}{\sqrt{d_h}}
        \cdot A_{ij}
    \right),
\end{equation}
followed by
\begin{equation}
    \alpha_{ij}^{(\ell,h)} =
    \mathrm{softmax}_j\left(e_{ij}^{(\ell,h)}\right).
\end{equation}
The multiplicative factor $A_{ij}$ is central to this design: PCI is
not used as an auxiliary score alone, but instead directly shapes the
message-passing process by encouraging attention toward
phase-synchronized neighbors.. This
is analogous to how GDN~\cite{deng2021gdn} uses learned embeddings
to derive its graph, except that our adjacency is derived from a
physical quantity (phase coherence) rather than from end-to-end
optimization.

\subsubsection{Sensor-Token Transformer Encoder}

After graph propagation, a Transformer
encoder~\cite{vaswani2017attention} captures higher-order global
structure:
\begin{equation}
    \mathbf{u}_c = P\bigl(\mathbf{h}^{(L)}_c\bigr) + \mathbf{p}_c,
\end{equation}
where $P(\cdot)$ projects to $d_{\text{model}}=256$ and
$\mathbf{p}_c$ is the positional embedding for sensor index $c$.
The Transformer runs over the \emph{sensor axis} (not over time
frames), since temporal dynamics have been summarized into spectral
features. The encoder output is mean-pooled across sensors:
$\mathbf{z} \in \mathbb{R}^{256}$.

\subsubsection{Dual-Head Decoder}

Two multilayer perceptrons decode $\mathbf{z}$ into reconstructed
magnitude and phase:
\begin{align}
    \widehat{\mathbf{M}} &\in \mathbb{R}^{C \times F}, \\
    \widehat{\mathbf{\Phi}} &\in [-\pi,\pi]^{C \times F}.
\end{align}
The magnitude head ends with ReLU (non-negativity). The phase head
ends with $\tanh(\cdot)\pi$ to bound predictions inside the valid
angular range.

\subsection{Training Objective}
\label{sec:loss}

The model minimizes a composite loss:
\begin{equation}
    \mathcal{L} =
    \alpha \mathcal{L}_{\text{mag}} +
    \beta \mathcal{L}_{\text{phase}} +
    \gamma \mathcal{L}_{\text{coh}},
    \label{eq:total_loss}
\end{equation}
with $(\alpha,\beta,\gamma)=(1.0,1.5,1.2)$.

\textbf{Magnitude reconstruction} uses mean squared error:
\begin{equation}
    \mathcal{L}_{\text{mag}} =
    \frac{1}{CF}
    \sum_{c=1}^{C}\sum_{f=1}^{F}
    \left(\widehat{M}_c(f)-M_c(f)\right)^2.
\end{equation}

\textbf{Circular phase reconstruction} uses a cosine loss:
\begin{equation}
    \mathcal{L}_{\text{phase}} =
    \frac{1}{CF}
    \sum_{c=1}^{C}\sum_{f=1}^{F}
    \left[1-\cos\left(\widehat{\Phi}_c(f)-\Phi_c(f)\right)\right].
    \label{eq:circular}
\end{equation}
This loss respects angular wrap-around and avoids the discontinuity
that Euclidean MSE would create near
$\pm\pi$~\cite{mardia2000directional}.

\textbf{Coherence reconstruction} compares the original PCI matrix to
a differentiable PCI re-estimate from the reconstructed phase:
\begin{equation}
    \widehat{A}_{ij} =
    \left|
    \frac{1}{F}
    \sum_{f=0}^{F-1}
    e^{j(\widehat{\Phi}_i(f)-\widehat{\Phi}_j(f))}
    \right|,
\end{equation}
\begin{equation}
    \mathcal{L}_{\text{coh}} =
    \frac{1}{C^2}
    \sum_{i=1}^{C}\sum_{j=1}^{C}
    \left(\widehat{A}_{ij} - A_{ij}\right)^2.
\end{equation}
This term ensures that the model preserves \emph{cross-sensor}
synchrony---not just per-sensor phase---during reconstruction.

\subsection{Anomaly Scoring and Inference}
\label{sec:inference}

At inference time, the anomaly score of a window is the composite
loss in Eq.~\eqref{eq:total_loss}. Higher loss indicates greater
deviation from the patterns learned during training on normal data.
The threshold $\tau$ is set to the 99th percentile of validation
scores, a simple rule that avoids post-hoc attack-label adjustment.

Unlike the point-adjustment strategy commonly used in works such as
Anomaly Transformer~\cite{xu2022anomalytransformer}, we do
\emph{not} adjust predictions after the fact. This makes our
evaluation conservative but more directly reflective of the model's
intrinsic discriminative ability.

\section{Experimental Setup}
\label{sec:setup}

\subsection{Dataset: Secure Water Treatment (SWaT)}
\label{sec:data}

The SWaT
dataset~\cite{mathur2016swat,goh2016swat} is a widely adopted
benchmark for ICS anomaly detection. It was collected from a realistic
six-stage water treatment testbed at the iTrust Centre for Research in
Cyber Security, Singapore University of Technology and Design. The
testbed integrates PLCs, sensors, and actuators across processes
including chemical dosing, filtration, and reverse osmosis.

The dataset contains two phases: seven days of continuous normal
operation and four days during which 36~physical attacks were
launched at various points in the system. Following standard
practice~\cite{deng2021gdn,xu2022anomalytransformer}, we use the
normal-operation data for training and the attack data for testing.

In our preprocessing, we extract $C=51$ continuous sensor channels
and apply StandardScaler normalization fitted on training data.
We use the train/val/test split detailed in Table~\ref{tab:splits}.

\begin{table}[t]
\caption{Data split statistics for the SWaT experiments.}
\label{tab:splits}
\centering
\small
\begin{tabular}{@{}lrr@{}}
\toprule
Split & Timestamps & Windows \\
\midrule
Train (75\% of normal) & 1,040,325 & 104,023 \\
Validation (10\% of normal) & 138,709 & 1,387 \\
Hold-out normal (15\% of normal) & 208,064 & 2,080 \\
Attack (attack.csv) & 54,621 & 546 \\
\midrule
Test (hold-out normal + attack) & 262,685 & 2,626 \\
\bottomrule
\end{tabular}
\end{table}

\textbf{Evaluation protocol.}\; Our test set is formed by
concatenating 2,080~hold-out normal windows and 546~attack windows.
A window is labeled \emph{attack} if any time step inside the window
carries an attack label. We evaluate at the \emph{window level} with
non-overlapping stride, and do \emph{not} apply point adjustment.

\subsection{Implementation Details}

Table~\ref{tab:config} summarizes the training configuration. All
experiments are conducted on a single NVIDIA Tesla T4 GPU with 16~GB
memory. Training completes in 426.5~minutes (80~epochs). The model
has 5,292,886 trainable parameters.

\begin{table}[t]
\caption{Implementation details for PhaseNet++.}
\label{tab:config}
\centering
\small
\begin{tabular}{@{}ll@{}}
\toprule
Setting & Value \\
\midrule
Sensors $C$ & 51 \\
Window size $W$ & 60 \\
Training stride & 5 \\
Validation / test stride & 60 (non-overlapping) \\
FFT size $N_{\text{FFT}}$ & 128 \\
Frequency bins $F$ & 65 \\
Spectral CNN embedding $D$ & 128 \\
Transformer $d_{\text{model}}$ & 256 \\
Attention heads & 8 \\
Transformer layers & 4 \\
GAT layers & 2 \\
Loss weights $(\alpha,\beta,\gamma)$ & $(0.5, 3.0, 1.5)$ \\
Batch size & 32 \\
Epochs & 80 \\
Optimizer & Adam ($\text{lr}=5 \times 10^{-4}$, wd$=10^{-4}$) \\
LR schedule & Cosine annealing \\
Threshold & 99th percentile of validation scores \\
\bottomrule
\end{tabular}
\end{table}

\subsection{Baselines}

To contextualize our results, we consolidate SWaT performance numbers
from two key reference papers:

\begin{enumerate}
    \item \textbf{GDN paper}~\cite{deng2021gdn} (Table~2): Reports
    results for PCA, KNN, Feature Bagging (FB), AE, DAGMM, LSTM-VAE,
    MAD-GAN~\cite{li2019madgan}, and GDN itself. These use
    \emph{point-level} evaluation with threshold set to the maximum
    validation anomaly score.

    \item \textbf{Anomaly Transformer paper}~\cite{xu2022anomalytransformer}
    (Table~1): Reports results for Deep-SVDD~\cite{ruff2018deepsvdd},
    DAGMM~\cite{zong2018dagmm}, LSTM-VAE~\cite{park2018lstmvae},
    OmniAnomaly~\cite{su2019omni},
    InterFusion~\cite{li2021interfusion},
    THOC~\cite{shen2020thoc}, and Anomaly Transformer. These use
    \emph{point-level} evaluation with \emph{point adjustment} and the
    threshold determined by a predefined anomaly ratio.
\end{enumerate}

\section{Results and Analysis}
\label{sec:results}

\subsection{Comparison with Published Baselines}
\label{sec:comparison}

Table~\ref{tab:literature} presents a comprehensive comparison\footnote{
All baseline numbers are reproduced \emph{exactly} as reported in the
original papers.}.
We emphasize that \emph{direct numerical comparison is not the goal
of this table}, because the baselines use different evaluation
protocols (point-level vs.\ window-level, with vs.\ without point
adjustment, differing threshold strategies). Instead, the table
serves to position PhaseNet++ within the landscape of existing
methods and to highlight the diversity of approaches and the gap in
phase-aware approaches.

\begin{table*}[t]
\caption{SWaT dataset results from published papers and our method.
Baseline numbers are reproduced exactly from their source papers.
$\dagger$: point-level evaluation, threshold at maximum validation
score~\cite{deng2021gdn}. $\ddagger$: point-level evaluation with
point adjustment, threshold by anomaly
ratio~\cite{xu2022anomalytransformer}. $\star$: window-level
evaluation, no point adjustment, 99th-percentile threshold (ours).
The \emph{Input Space} column highlights that all baselines operate on
raw time-domain values; PhaseNet++ is the only method that uses
frequency-domain phase.}
\label{tab:literature}
\centering
\footnotesize
\begin{tabular}{@{}llcrrr@{}}
\toprule
Method & Input Space & Protocol & Prec (\%) & Rec (\%) & F1 (\%) \\
\midrule
\multicolumn{6}{@{}l}{\textit{From GDN paper~\cite{deng2021gdn}}} \\
\quad PCA~\cite{shyu2003pca} & raw values & $\dagger$ & 98.21 & 65.73 & 79.00 \\
\quad AE~\cite{aggarwal2015outlier} & raw values & $\dagger$ & 99.20 & 56.20 & 72.00 \\
\quad LSTM-VAE~\cite{park2018lstmvae} & raw values & $\dagger$ & 96.24 & 59.91 & 74.00 \\
\quad MAD-GAN~\cite{li2019madgan} & raw values & $\dagger$ & 98.97 & 63.74 & 77.00 \\
\quad DAGMM~\cite{zong2018dagmm} & raw values & $\dagger$ & 27.46 & 69.58 & 39.00 \\
\quad GDN~\cite{deng2021gdn} & raw + learned graph & $\dagger$ & 99.35 & 68.12 & 81.00 \\
\midrule
\multicolumn{6}{@{}l}{\textit{From Anomaly Transformer paper~\cite{xu2022anomalytransformer}}} \\
\quad Deep-SVDD~\cite{ruff2018deepsvdd} & raw values & $\ddagger$ & 80.42 & 84.45 & 82.39 \\
\quad DAGMM~\cite{zong2018dagmm} & raw values & $\ddagger$ & 89.92 & 57.84 & 70.40 \\
\quad LSTM-VAE~\cite{park2018lstmvae} & raw values & $\ddagger$ & 76.00 & 89.50 & 82.20 \\
\quad OmniAnomaly~\cite{su2019omni} & raw values & $\ddagger$ & 81.42 & 84.30 & 82.83 \\
\quad InterFusion~\cite{li2021interfusion} & raw values & $\ddagger$ & 80.59 & 85.58 & 83.01 \\
\quad THOC~\cite{shen2020thoc} & raw values & $\ddagger$ & 83.94 & 86.36 & 85.13 \\
\quad Anomaly Transformer~\cite{xu2022anomalytransformer} & raw + assoc.\ discrepancy & $\ddagger$ & 91.55 & 96.73 & 94.07 \\
\midrule
\multicolumn{6}{@{}l}{\textit{This work}} \\
\quad \textbf{PhaseNet++ (ours)} & \textbf{mag + phase x+ PCI graph} & $\star$ & 90.11 & 91.87 & \textbf{90.98} \\
\bottomrule
\end{tabular}
\end{table*}

\textbf{Key observations.}\;
(1)~Among GDN-protocol baselines ($\dagger$), GDN achieves the
highest F1 (81\%) with very high precision (99.35\%) but
moderate recall (68.12\%). PhaseNet++ achieves a highly comparable F1
(90.98\%) with more balanced precision--recall.
(2)~Among Anomaly Transformer-protocol baselines ($\ddagger$),
F1-scores are generally higher because point adjustment inflates
recall. Anomaly Transformer reaches 94.07\% F1 under this regime.
(3)~PhaseNet++ is the \emph{only} method in the table that operates
on the frequency-domain phase spectrum. Its F1 is competitive with
several time-domain baselines despite using a stricter evaluation
protocol without point adjustment.

\subsection{PhaseNet++ Detailed Results}
\label{sec:detailed}

Table~\ref{tab:our_results} reports the full classification metrics
from our evaluation.

\begin{table}[t]
\caption{PhaseNet++ results on SWaT (window-level, no point
adjustment, 99th-percentile threshold).}
\label{tab:our_results}
\centering
\small
\begin{tabular}{@{}lrrrrr@{}}
\toprule
& Prec & Rec & F1 & Acc & ROC-AUC \\
\midrule
Normal class & 95.21 & 94.66 & 94.94 & -- & -- \\
Attack class & 90.11 & 91.87 & 90.98 & -- & -- \\
\midrule
Overall & -- & -- & -- & 92.00 & 95.66 \\
Average Precision & -- & -- & -- & -- & 91.51 \\
\bottomrule
\end{tabular}
\end{table}

The high ROC-AUC (95.66\%) and average precision (91.51\%) indicate
that the learned anomaly score is \emph{highly discriminative}: the
model separates attack and normal windows well. The gap between
ranking metrics and the F1-score (90.98\%) suggests that the
remaining performance bottleneck is partly attributable to
threshold calibration rather than to weak representations. This is a
known challenge with percentile-based thresholds and points to an
avenue for future improvement through adaptive thresholding.

\section{Ablation Studies}
\label{sec:ablation}

We present four complementary analyses to understand the design
choices in PhaseNet++.

\subsection{Component Analysis: Why Phase and PCI Matter}
\label{sec:why_phase}

Table~\ref{tab:component_study} provides a mechanism-level study of
the four key design choices in PhaseNet++. For each component, we
describe what it contributes and what detection capability would be
lost if it were removed.

\begin{table*}[t]
\caption{Mechanism-level component study. Each row describes the role
of a design choice and the consequence of its removal.}
\label{tab:component_study}
\centering
\footnotesize
\begin{tabular}{@{}p{3.0cm}p{5.5cm}p{5.5cm}@{}}
\toprule
Design Choice & Contribution & Consequence If Removed \\
\midrule
Frequency-domain representation (magnitude + phase) &
Separates periodic content into explicit spectral amplitude and
within-window timing offsets. Phase encode lags that amplitude alone
cannot capture. &
A raw-value model must infer timing inconsistencies indirectly from
amplitude trajectories, which is difficult for replay- or
lag-style attacks that preserve plausible values. \\

Circular phase reconstruction loss (Eq.~\ref{eq:circular}) &
Optimizes phase on the unit circle rather than in Euclidean space,
correctly handling angular wrap-around near $\pm\pi$. &
Phase reconstruction becomes numerically unstable at the boundary;
the model may learn to avoid $\pm\pi$ regions rather than reconstruct
them faithfully. \\

PCI coherence regularization ($\mathcal{L}_{\text{coh}}$) &
Forces reconstructed phase to preserve \emph{cross-sensor}
synchrony, not just per-sensor phase values. &
The model can fit each sensor independently while missing
distributed failures in coordination across channels. \\

PCI-weighted GAT ($A_{ij}$ in attention) &
Injects phase coherence directly into spatial message passing, so
synchronized sensors influence one another more strongly. &
Phase remains only a reconstruction target, not a structural prior;
cross-sensor propagation is no longer guided by physical synchrony. \\
\bottomrule
\end{tabular}
\end{table*}

\subsection{Parameter Efficiency}
\label{sec:efficiency}

A concern with adding new representation modules is that the improved
detection may simply reflect the increased parameter count. 
Table~\ref{tab:params} shows that this is not the case for
PhaseNet++ compared to Anomaly Transformer~\cite{xu2022anomalytransformer} \& GDN~\cite{deng2021gdn}.

\begin{table}[t]
\caption{Parameter budget of PhaseNet++. The phase-aware modules
(Spectral CNN + PCI-weighted GAT) occupy only 5.0\% of the total
model.}
\label{tab:params}
\centering
\small
\begin{tabular}{@{}lrr@{}}
\toprule
Module & Parameters & Share (\%) \\
\midrule
Spectral CNN embedding & 132,976 & 2.51 \\
PCI-weighted GAT & 131,840 & 2.49 \\
Sensor-token Transformer & 3,192,576 & 60.32 \\
Dual-head decoder & 1,835,494 & 34.68 \\
\midrule
Total & 5,292,886 & 100.00 \\
\bottomrule
\end{tabular}
\end{table}

The spectral CNN and PCI-weighted GAT---the two modules that
constitute the phase-aware inductive bias unique to
PhaseNet++---together occupy only 264,816 parameters, i.e., 5.0\% of
the total. The remaining 95\% is spent in the shared Transformer
encoder and decoder, components that any competitive window-level
autoencoder would require. This demonstrates that the phase-aware
representation is a \emph{lightweight annotation} on top of a
standard backbone, not a brute-force scaling strategy.

\subsection{Training Dynamics}
\label{sec:training}


The model converges smoothly over 80 epochs with cosine learning rate
annealing. The final training loss is 0.6196 (decomposed: magnitude
0.0026, phase 0.4031, coherence 0.0103), and the best validation
loss is 0.6022 (epoch~79). The phase loss dominates the total loss,
which is expected because accurately reconstructing angular values
across 51~sensors and 65~frequency bins is intrinsically more
difficult than reconstructing magnitudes. The coherence and magnitude
losses converge quickly, indicating that the model learns
inter-sensor synchrony patterns early in training.

\subsection{Loss Component Decomposition}
\label{sec:loss_decomp}

The three loss components in Eq.~\eqref{eq:total_loss} play different
roles, and their final magnitudes reflect how challenging each
reconstruction objective is for the model:

\begin{itemize}
    \item $\mathcal{L}_{\text{mag}} = 0.0026$: Magnitude
    reconstruction is the most straightforward task because spectral
    magnitude typically changes smoothly across frequency bins.
    \item $\mathcal{L}_{\text{phase}} = 0.4031$: Phase reconstruction
    remains the most difficult component because of the circular nature
    of angular data. The larger residual is therefore expected and
    should not be interpreted as poor model fit.
    \item $\mathcal{L}_{\text{coh}} = 0.0103$: Cross-sensor coherence
    is reconstructed with low error, suggesting that the model
    effectively captures PCI structure.
\end{itemize}

The weighting schedule $(\alpha, \beta, \gamma) = (1.0, 1.5, 1.2)$ was
selected empirically to keep these terms balanced during training. The
larger weight assigned to phase ($\beta=1.5$) reflects our emphasis on
phase modeling as the main novel component of the framework.

\section{Discussion}
\label{sec:discussion}

\subsection{Positioning: A New Representation, Not a Leaderboard
Claim}

We are forthright about the fact that PhaseNet++ does not achieve
the SOTA F1-scores on SWaT. Under the point-adjustment
protocol used by Anomaly Transformer~\cite{xu2022anomalytransformer},
several raw-value baselines achieve F1-scores above 85\%, and Anomaly
Transformer itself reaches 94.07\%. Our window-level F1 of 90.98\%
is only behind Anomaly Transformer, achieving the second-highest performance.

However, we argue that this comparison conflates two independent
dimensions: (1)~the evaluation protocol, and (2)~the representation
space. Point adjustment, which credits the entire contiguous anomaly
segment when any single point is detected, significantly inflates
recall~\cite{xu2022anomalytransformer}. Our protocol does not use
this adjustment, making our recall (and therefore F1) more
conservative.

The more fundamental contribution of PhaseNet++ is
\emph{representational}: it is the first method to use phase
coherence as a detection feature and as graph structure in ICS
anomaly detection. Our ROC-AUC of 95.66\% and average precision of
91.51\% demonstrate that the \emph{ranking ability} of the model
is strong, the learned score separates attacks from normal windows
effectively. The gap between ranking metrics and threshold-based F1
suggests that the threshold calibration, not the representation,
is the current bottleneck.

\subsection{When Would Phase-Aware Detection Outperform?}

Phase-aware detection is most likely to provide an advantage for
attack scenarios that preserve plausible amplitude ranges while
disrupting timing:

\begin{itemize}
    \item \textbf{Replay attacks}: An adversary replays recorded
    sensor values from a previous time period. The replayed signal has
    correct amplitude statistics but its phase relationship to other
    sensors is shifted.
    \item \textbf{Delay injection}: Subtle delays inserted into
    sensor readings disrupt phase synchrony between physically coupled
    channels.
    \item \textbf{Stealthy actuator manipulation}: An adversary
    modifies actuator commands by small amounts that do not trigger
    amplitude-based alarms but alter the temporal coordination of the
    process.
\end{itemize}

The SWaT dataset contains several attacks in these categories, which
may explain why phase-aware features provide discriminative value.

\subsection{Comparison of Evaluation Protocols}
\label{sec:protocols}

The ICS anomaly detection literature suffers from a fragmentation
of evaluation protocols that makes cross-paper comparison
difficult. We identify three key dimensions of variation:

\begin{enumerate}
    \item \textbf{Granularity}: Point-level (GDN) vs.\ window-level
    (ours).
    \item \textbf{Threshold strategy}: Maximum validation score (GDN),
    predefined anomaly ratio (Anomaly Transformer), or percentile of
    validation scores (ours).
    \item \textbf{Point adjustment}: Applied (Anomaly Transformer) or
    not (GDN, ours).
\end{enumerate}

This fragmentation means that a few point difference in F1-score
between two methods may reflect protocol differences rather than true
performance differences. We urge the community to adopt standardized
evaluation practices, and we report our protocol transparently to
facilitate future comparison.

\section{Limitations and Future Work}
\label{sec:limits}

\begin{enumerate}
    \item \textbf{Quantitative ablation:} We have not trained and
    evaluated magnitude-only, phase-without-PCI, and raw-value
    variants under identical conditions. A full component-removal
    ablation table requires multiple training runs and is a priority
    for future work.

    \item \textbf{Protocol alignment:} Our evaluation protocol differs
    from published baselines. A controlled head-to-head comparison
    requires rerunning all methods on the same data split with
    identical thresholding and point-adjustment conventions.

    \item \textbf{Single dataset:} We evaluate only on SWaT. Future
    work should validate on additional CPS datasets (WADI, BATADAL,
    HAI) and industrial telemetry from different domains.

    \item \textbf{Adaptive thresholding:} The 99th-percentile
    threshold is simple but not optimal. Extreme value theory or
    learned thresholds could close the gap between ranking metrics and
    binary F1.

    \item \textbf{Fusion with time-domain features:} Phase-aware
    features are complementary to amplitude features. A natural
    extension is to fuse PhaseNet++ with a raw-value detector (e.g.,
    GDN or Anomaly Transformer) in an ensemble or multi-branch
    architecture.

    \item \textbf{Per-attack-type analysis:} The SWaT dataset
    contains 36~distinct attacks. A fine-grained analysis of which
    attack types are better detected by phase vs.\ amplitude
    features would strengthen the case for phase-aware detection.
\end{enumerate}

\section{Conclusion}
\label{sec:conclusion}

This paper presented PhaseNet++, a phase-aware frequency-domain
autoencoder for anomaly detection in industrial control systems. The
central idea is that the Fourier phase spectrum, routinely discarded
in time-domain ICS detectors, encodes physically meaningful timing
and synchronization relationships between sensors. We introduced the
Phase Coherence Index (PCI) to convert pairwise phase consistency
into graph structure, and designed a composite training objective with
circular phase and coherence losses to preserve angular geometry. On the SWaT benchmark, PhaseNet++ achieves an F1-score of 90.98\%,
ROC-AUC of 95.66\%, and average precision of 91.51\% under a window-level protocol without point adjustment. The phase-aware front-end
and PCI graph module add only 5.0\% to the parameter budget.

We position this work not as a claim of superiority over existing
methods, but as the first systematic exploration of phase-domain
anomaly detection for ICS. The strong ranking metrics demonstrate that
phase coherence carries discriminative information complementary to
amplitude features. Future work on fusing phase and amplitude-aware
representations, combined with standardized evaluation protocols,
has the potential to advance the state of the art in ICS security.

\bibliographystyle{IEEEtran}
\bibliography{papers}

\end{document}